\documentclass[]{memtensor}

\usepackage[toc,page,header]{appendix}
\usepackage[utf8]{inputenc} 
\usepackage[T1]{fontenc}    
\usepackage{hyperref}       
\usepackage{url}            
\usepackage{booktabs}       
\usepackage{amsfonts}       
\usepackage{nicefrac}       
\usepackage{microtype}      
\usepackage{xcolor}         
\usepackage{amsmath} 
\usepackage{etoolbox}
\usepackage{lipsum}  
\usepackage{minitoc}
\usepackage[table]{xcolor}  
\usepackage{tablefootnote}  
\usepackage{threeparttable}

\title{MemFactory: Unified Inference \& Training Framework for Agent Memory}

\author{Ziliang Guo}
\author{Ziheng Li}
\author{Bo Tang}
\author{Feiyu Xiong}
\author[\dagger]{Zhiyu Li}

\affiliation[]{MemTensor}

\abstract{
Memory-augmented Large Language Models (LLMs) are essential for developing capable, long-term AI agents. Recently, applying Reinforcement Learning (RL) to optimize memory operations—such as extraction, updating, and retrieval—has emerged as a highly promising research direction. However, existing implementations remain highly fragmented and task-specific, lacking a unified infrastructure to streamline the integration, training, and evaluation of these complex pipelines. To address this gap, we present \textbf{MemFactory}, the first unified, highly modular training and inference framework specifically designed for memory-augmented agents. Inspired by the success of unified fine-tuning frameworks like LLaMA-Factory, MemFactory abstracts the memory lifecycle into atomic, plug-and-play components, enabling researchers to seamlessly construct custom memory agents via a ``Lego-like'' architecture. Furthermore, the framework natively integrates Group Relative Policy Optimization (GRPO) to fine-tune internal memory management policies driven by multi-dimensional environmental rewards. MemFactory provides out-of-the-box support for recent cutting-edge paradigms, including Memory-R1, RMM, and MemAgent. We empirically validate MemFactory on the open-source MemAgent architecture using its publicly available training and evaluation data. Across the evaluation sets, MemFactory improves performance over the corresponding base models on average, with relative gains of up to 14.8\%. By providing a standardized, extensible, and easy-to-use infrastructure, MemFactory significantly lowers the barrier to entry, paving the way for future innovations in memory-driven AI agents.
}
\date{\today}
\correspondence{Team Leader at \email{lizy@memtensor.cn}}
\checkdata[Author Legend]{\dag{}Corresponding Author}
\checkdata[Code]{\url{https://github.com/MemTensor/MemFactory}}

\begin{document}
\maketitle

\section{Introduction}

Large Language Models (LLMs) have demonstrated remarkable capabilities in natural language understanding, reasoning, and generation. However, to evolve from simple conversational interfaces into fully autonomous AI agents, these models must possess the ability to maintain long-term context, accumulate historical experiences, and personalize their behaviors over continuous interactions. Consequently, memory-augmented LLMs have become a critical focal point in current AI research. While early approaches predominantly relied on static Retrieval-Augmented Generation (RAG) paradigms or heuristic-based memory updating rules, these methods often struggle with memory redundancy, conflicting information, and context-limit constraints during prolonged execution. 

To overcome these limitations, a highly promising trend has emerged: optimizing memory operations through Reinforcement Learning (RL). By treating memory management as a sequential decision-making process, RL allows agents to learn optimal policies for \textit{when} and \textit{what} to extract, update, or retrieve based on delayed environmental feedback. Recent pioneering works, such as Memory-R1 \cite{yan2026memoryr1}, MemAgent \cite{yu2025memagent}, and RMM \cite{tan2025prospectretrospectreflectivememory}, have successfully demonstrated that models trained with RL can autonomously develop sophisticated memory strategies, significantly outperforming their heuristically driven counterparts.

Despite these rapid advancements, the ecosystem for Memory-RL research remains heavily fragmented. Current implementations are typically highly customized, deeply coupled with specific tasks, and scattered across isolated repositories. For researchers and developers, reproducing these memory strategies, or combining different modules (e.g., swapping a retrieval module in Memory-R1 with an LRM-based reranker) requires non-trivial and redundant engineering efforts. In the realm of standard LLM fine-tuning, unified frameworks like LLaMA-Factory \cite{zheng2024llamafactory} have successfully democratized model adaptation by providing systematic, code-free, and highly scalable infrastructures. Yet, a comparable unified infrastructure specifically tailored for the complex lifecycle of memory-augmented agents remains glaringly absent.

To bridge this crucial gap, we introduce \textbf{MemFactory}, an open-source, modular, and unified training and inference framework designed to streamline the development of memory-augmented agents. Inspired by the architectural elegance of LLaMA-Factory \cite{zheng2024llamafactory}, MemFactory abstracts the intricate memory pipeline into standardized, plug-and-play atomic operations—namely extraction, updating, and retrieval. By orchestrating these atomic modules through a flexible \textit{Agent Layer}, researchers can effortlessly construct, customize, and experiment with diverse memory architectures. Crucially, MemFactory is built from the ground up to support RL-driven policy optimization. Through its integrated \textit{Trainer Layer}, the framework natively employs Group Relative Policy Optimization (GRPO) to fine-tune the agent's internal memory management strategies based on multi-dimensional rewards from the environment.

In summary, the main contributions of our work are as follows:
\begin{itemize}
    \item \textbf{A Unified Memory-RL Infrastructure:} We present MemFactory, the first comprehensive framework that unifies the training, evaluation, and inference pipelines for memory-augmented AI agents, significantly lowering the engineering barrier for future research.
    \item \textbf{Highly Modular and Extensible Design:} By decoupling the memory lifecycle into atomic modules (Extractors, Updaters, Retrievers), MemFactory enables a ``Lego-like'' assembly paradigm, allowing seamless integration and mixing of various memory strategies.
    \item \textbf{Out-of-the-box Support for SOTA Paradigms:} The framework natively incorporates and standardizes recent cutting-edge Memory-RL works, including Memory-R1, MemAgent, and RMM, providing readily executable baselines.
    \item \textbf{Empirical Validation:} We empirically demonstrate the effectiveness of MemFactory by training an open-source MemAgent architecture using our built-in GRPO pipeline, achieving consistent improvements on the main-task evaluation sets and overall gains across the full evaluation suite.
\end{itemize}

\section{Related Work}

\subsection{Unified LLM Training and Inference Frameworks}

The proliferation of Large Language Models (LLMs) has catalyzed the development of standardized infrastructures for efficient model adaptation. Beyond the foundational capabilities of LLaMA-Factory \cite{zheng2024llamafactory}, which unifies diverse Parameter-Efficient Fine-Tuning (PEFT) and alignment algorithms (e.g., SFT, DPO) within a modular architecture, other integrated frameworks have emerged to address specific scaling and alignment needs. For instance, MS-Swift \cite{zhao2025swift} provides a versatile ecosystem for multi-modal model fine-tuning and deployment, while specialized libraries like VERL \cite{Sheng_2025} offer high-throughput infrastructures specifically optimized for large-scale Reinforcement Learning (RL) pipelines. 

Despite the success of these frameworks in standardizing LLM alignment, they remain predominantly designed for stateless sequence-to-sequence modeling. They lack the architectural primitives required to manage the complex, stateful lifecycle of memory-augmented AI agents, such as iterative memory extraction, dynamic database updating, or sequential interactions with evolving environments. Recent memory-centric systems have begun to standardize the construction of memory modules themselves. For example, MemEngine \cite{zhang2025memengine} provides a unified and modular library to develop LLM-based memory agents, while Mem0 \cite{chhikara2025mem0buildingproductionreadyai} offers a scalable memory-centric architecture for extracting, consolidating, and retrieving long-term conversational memories. MemFactory draws architectural inspiration from the ``Lego-like'' modularity of LLaMA-Factory \cite{zheng2024llamafactory} and the RL-centric design of VERL, while also taking cues from recent memory-centric abstractions that improve the composability of memory modules. In this way, MemFactory fundamentally re-engineers the infrastructure to support the unique paradigm of Memory-RL. By integrating static fine-tuning, modular memory construction, and dynamic policy optimization, the framework enables researchers to easily build, train, and evaluate customized memory agents.

\subsection{Adaptive Memory via Reinforcement Learning}

Traditional memory-augmented LLMs typically rely on heuristic rules and static pipelines for memory storage and retrieval. These pre-defined strategies struggle with long-horizon tasks, frequently accumulating noise or misinterpreting conflicting information. To overcome these limitations, recent research frames memory management as a sequential decision-making problem, utilizing Reinforcement Learning (RL) to dynamically optimize memory policies via outcome-based rewards. 

Pioneering works in this direction have explored distinct facets of the memory lifecycle:
\begin{itemize}
    \item \textbf{Structured Memory Operations:} \textit{Memory-R1} \cite{yan2026memoryr1} trains a Memory Manager to execute discrete CRUD operations (e.g., \texttt{ADD}, \texttt{UPDATE}, \texttt{DELETE}) and an Answer Agent for memory distillation. Optimized via PPO and GRPO, it demonstrates that exact-match rewards are sufficient to teach sophisticated memory consolidation without dense human annotation.
    \item \textbf{Recurrent State Optimization:} To bypass the quadratic bottlenecks of long contexts, \textit{MemAgent} \cite{yu2025memagent} treats a fixed-length latent memory variable as a recurrent state across text segments. It employs Multi-Conversation DAPO to learn an optimal ``overwrite'' policy, aggressively compressing history while preserving task-critical facts.
    \item \textbf{Retrieval Refinement:} \textit{RMM} \cite{tan2025prospectretrospectreflectivememory} dynamically optimizes memory retrieval for personalized dialogues via ``Retrospective Reflection.'' It leverages unsupervised LLM attribution signals (i.e., whether a retrieved memory was actually cited) as binary rewards to update a lightweight reranker online via the REINFORCE algorithm.
\end{itemize}

Despite demonstrating the efficacy of RL across diverse memory paradigms—from explicit database updating to recurrent state compression and retrieval optimization—these implementations remain heavily siloed. Each is tightly coupled to custom training pipelines, data formats, and task assumptions. \textbf{MemFactory} resolves this fragmentation. By abstracting these core mechanisms into unified modules (e.g., the \texttt{Updater} for Memory-R1, the \texttt{Agent Module} for MemAgent, and the \texttt{RerankRetriever} for RMM), our framework provides a consolidated platform to seamlessly reproduce, evaluate, and integrate these state-of-the-art algorithms out-of-the-box.

\subsection{Reinforcement Learning for Policy Optimization}

Reinforcement Learning (RL) has become central to aligning Large Language Models (LLMs) with complex reasoning tasks \cite{ouyang2022traininglanguagemodelsfollow}. While Proximal Policy Optimization (PPO) \cite{schulman2017proximalpolicyoptimizationalgorithms} is the prevailing algorithm, it requires an auxiliary value network (critic) comparable in size to the policy model. This effectively doubles the training memory footprint—a prohibitive overhead for memory-augmented agents, where context windows are already severely saturated by prolonged dialogue histories and retrieved memories.

To address this computational bottleneck, Group Relative Policy Optimization (GRPO) \cite{shao2024GRPOdeepseekmathpushinglimitsmathematical} entirely eliminates the need for a parameterized value model. For a given input, GRPO samples a group of $G$ candidate responses and estimates the baseline through intra-group reward normalization. Specifically, the advantage $\hat{A}_i$ for the $i$-th candidate with reward $r_i$ is calculated as:
\begin{equation}
    \hat{A}_i = \frac{r_i - \text{mean}(\{r_1, \dots, r_G\})}{\text{std}(\{r_1, \dots, r_G\})}
\end{equation}
This formulation significantly reduces memory requirements while maintaining high sample efficiency, particularly when optimizing rule-based or outcome-driven rewards (e.g., exact-match or LLM-as-a-judge scores) prevalent in memory evaluation tasks.

\textbf{MemFactory} natively integrates GRPO into its \textit{Trainer Layer}. By leveraging GRPO's memory efficiency, our framework substantially lowers the hardware barriers for Memory-RL research. It enables researchers to efficiently fine-tune complex, long-context memory policies—spanning extraction, updating, and recurrent state transitions—even under constrained computational resources.

\section{MemFactory Framework}

\begin{figure}[htbp]
    \centering
    \includegraphics[width=0.95\linewidth, trim={1cm 1cm 1cm 1cm}, clip]{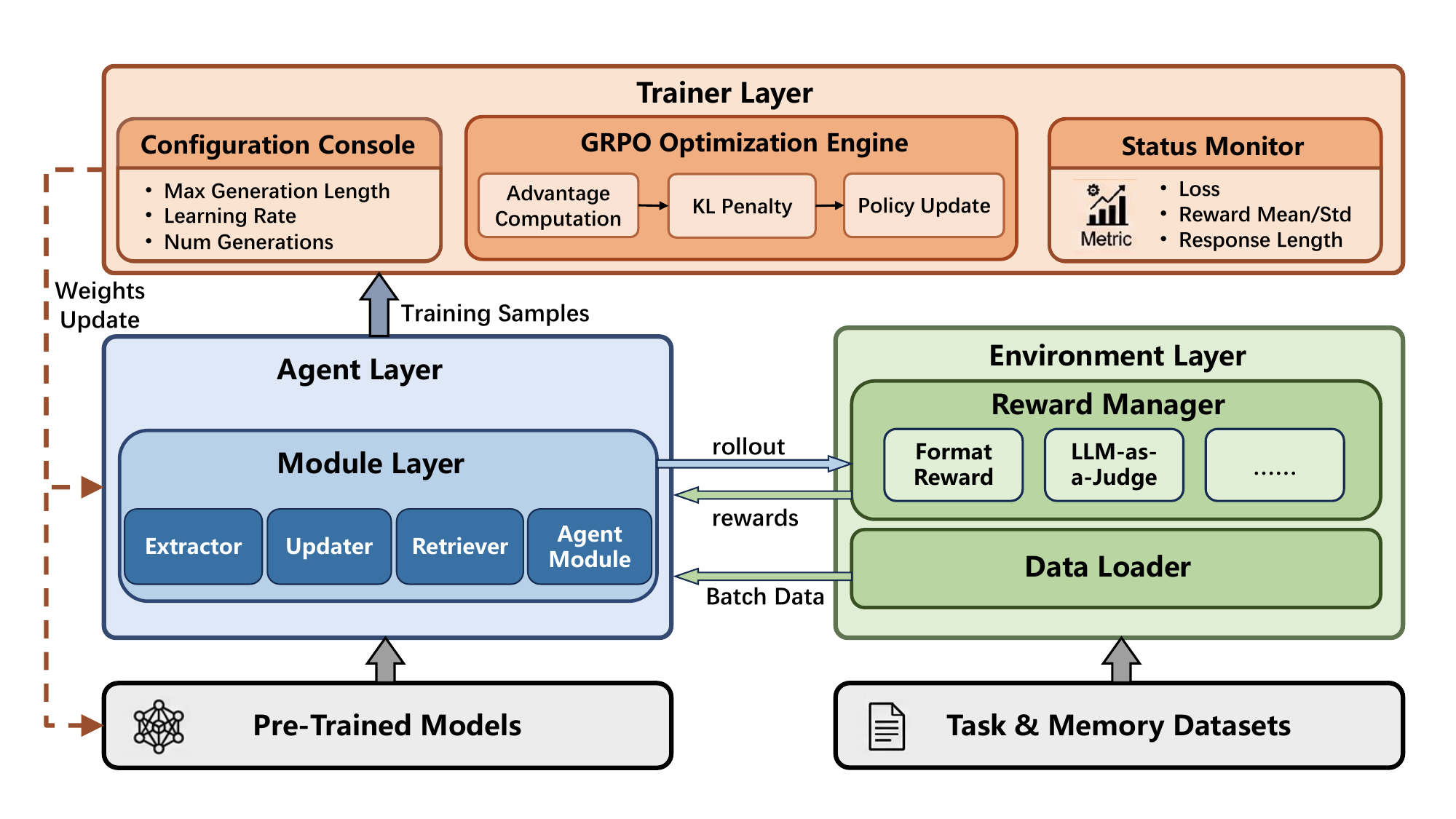}
    \caption{The overall architecture of MemFactory, illustrating the interdependencies among the Modules, Agent, Environment, and Trainer layers.}
    \label{fig:architecture}
\end{figure}

\textbf{MemFactory} is designed with a highly modular architecture comprising four primary components: \textit{Module Layer}, \textit{Agent Layer}, \textit{Environment Layer}, and \textit{Trainer Layer}. The \textit{Module Layer} operates as the fundamental core of the framework, abstracting the complex memory engineering pipeline into atomic operations, such as memory extraction, updating, and retrieval. By representing these atomic behaviors as plug-and-play modules, the framework ensures high extensibility and customization. The \textit{Agent Layer} builds upon the \textit{Module Layer} and serves as the central policy executor by systematically assembling these modules to implement various memory strategies and generate rollout trajectories during interaction. The \textit{Environment Layer} functions as both a dataloader and a reward manager. It standardizes raw data from diverse datasets into unified contextual states and evaluates the agent's actions to provide multi-dimensional reward signals. Finally, the \textit{Trainer Layer} employs Group Relative Policy Optimization (GRPO) to fine-tune the pre-trained models loaded within the agents, optimizing their internal memory management policies based on the environment's feedback. Figure \ref{fig:architecture} illustrates the overall architecture and the interdependencies of these components within MemFactory.

\subsection{\textit{Module Layer}: Atomic Memory Operations}

The \textit{Module Layer} is designed to decompose the complex memory lifecycle into manageable, atomic operations. Inspired by pioneering memory systems such as MemOS \cite{li2025memosmemoryosai} and Mem0 \cite{chhikara2025mem0buildingproductionreadyai}, we formulate the memory pipeline into three fundamental operations: extraction, update, and retrieval, along with their corresponding module classes. Furthermore, to accommodate end-to-end memory policies seen in works like MemAct \cite{zhang2026memact} and MemAgent \cite{yu2025memagent}, we introduce an additional \textit{Agent Module} class within this layer.

Each class of modules supports various internal implementations. However, they all conform to standardized interfaces to facilitate integration with the \textit{Agent Layer} and the \textit{Trainer Layer}. Specifically, a module typically implements three core methods: \texttt{generate}, \texttt{rollout}, and \texttt{inference}. Table \ref{tab:module_interfaces} outlines the specific roles of these methods.

\begin{table}[htbp]
    \centering
    \caption{Standard interfaces supported by modules in the \textit{Module Layer}.}
    \label{tab:module_interfaces}
    \resizebox{\textwidth}{!}{
    \begin{tabular}{llp{3.8cm}p{6cm}}
        \toprule
        \textbf{Interface} & \textbf{Model} & \textbf{Reward} & \textbf{Notes} \\
        \midrule
        \texttt{generate} & Training Model & Deferred & Produces intermediate states; reward is assigned only after the full interaction or trajectory is completed. \\
        \texttt{rollout} & Training Model & Final reward & Produces trajectories that receive the final task-level reward. \\
        \texttt{inference} & Inference Model / APIs & None & Used only for inference and is not trainable. \\
        \bottomrule
    \end{tabular}
    }
\end{table}

In the following, we introduce the functions, design, and implementation of these four modules.

\subsubsection{Memory Extractor}
The primary function of the memory extractor is to parse the raw contexts into structured memory pieces. It extracts facts, experiences, and insights from the historical context and saves them as discrete memory entries. Inspired by Memory-R1 \cite{yan2026memoryr1}, we implement the \texttt{NaiveExtractor} class to handle this extraction process.

\subsubsection{Memory Updater}
After the extractor generates candidate memory entries, the updater compares these candidates with existing memories to produce a memory update plan. Similarly, we draw inspiration from Memory-R1 \cite{yan2026memoryr1} to design the \texttt{NaiveUpdater} class, ensuring that the memory bank remains concise and highly accurate during prolonged user interactions. Specifically, the updater assigns one of four operations to manage the memory states: \texttt{ADD} to incorporate new information, \texttt{DEL} to remove obsolete or contradicted facts, \texttt{UPDATE} to modify existing entries, and \texttt{NONE} for memories that do not require changes.

\subsubsection{Memory Retriever}
The memory retriever is tasked with fetching the most relevant memory from the memory bank to ground the agent's responses. We implement the \texttt{NaiveRetriever} to perform standard semantic-based retrieval. To further enhance retrieval precision, we also provide the \texttt{RerankRetriever}. Reranking is a widely adopted post-retrieval technique \cite{hu2026memoryageaiagents}. In our implementation, the \texttt{RerankRetriever} leverages Large Reasoning Models (LRMs) to re-evaluate and rerank the initially retrieved memories, ultimately outputting the refined retrieval results.

\subsubsection{Agent Module}
For certain highly integrated memory strategies, strictly decoupling the memory pipeline into separate extraction and update phases is unnecessary. For example, in the paradigm proposed by MemAgent \cite{yu2025memagent}, the agent synthesizes new memory states directly from the latest context and previous memory states. When utilizing the memory, it places the entire memory directly into the context without performing any retrieval. To natively support such end-to-end approaches, we define the \textit{Agent Module} class. Thanks to the open-source release of MemAgent \cite{yu2025memagent}, we implemented the \texttt{RecurrentMemoryModule} following its design. This allows researchers to optimize unified, recurrent memory policies directly within our GRPO training framework.

\subsection{\textit{Agent Layer}: Module Integration and Policy Execution}

The \textit{Agent Layer} builds upon the \textit{Module layer} to serve as the central policy executor. We adopt a composable, ``Lego-like'' paradigm to construct robust memory agents. Researchers are highly encouraged to build custom agents tailored to specific tasks by swapping underlying module implementations and experimenting with various module interface combinations.

\subsubsection{Agent Initialization}
During initialization, we use \texttt{AutoClasses} from the Transformers library \cite{wolf2020transformers} to load pre-trained models. To efficiently handle long-context tasks and reduce memory usage, we integrate FlashAttention-2 \cite{dao2023flashattention2}. It should be noted that the loaded pre-trained model is the direct target for optimization. After the agent completes its rollout and environment interaction, the \textit{Trainer Layer} directly fine-tunes this specific model.

\subsubsection{Agent Rollout}
The agent's rollout process is implemented by its constituent modules using their designated interfaces. When a module is used purely for inference and not for training, the agent calls the \texttt{inference} interface. This interface supports OpenAI-style APIs and high-throughput inference engines such as vLLM \cite{kwon2023vllm}. 

Conversely, during the training phase, the agent uses the \texttt{generate} or \texttt{rollout} interfaces. To prevent tensor dimension mismatch during batch generation, we dynamically pad prompt tensors on the left and generated response tensors on the right. We also align the action masks with the sequence lengths to ensure accurate advantage computation in the subsequent reinforcement learning pipeline.

\subsubsection{Agent Inference}
Although MemFactory is primarily a training framework, it inherently supports pure inference. Researchers can instantiate an agent using the same modular composition, configuring each module to use the \texttt{inference} interface while bypassing the \textit{Trainer Layer}. This capability provides a practical way to compare different memory workflows and module combinations before committing to computationally expensive RL optimization.

\subsection{\textit{Environment Layer}: Data Processing and Reward Computation}

The \textit{Environment Layer} serves as the interface between the agent and the task. It processes raw datasets into standardized states and computes reward signals for policy optimization. To address different application scenarios, we draw inspiration from the ``long-term vs. short-term memory'' taxonomy proposed in the survey \textit{Memory in the Age of AI Agents} \cite{hu2026memoryageaiagents}. Based on this, we design two primary environments: \texttt{MemoryBankEnv} for long-term memory management, and \texttt{LongcontextEnv} for short-term, context-based memory.

As a dataloader, this layer converts diverse raw data into unified dictionary states. The main difference between the two environments is how they handle context: \texttt{MemoryBankEnv} maintains an explicit, updatable memory bank, whereas \texttt{LongcontextEnv} directly processes prolonged dialogue histories. As a reward manager, the environment provides multi-dimensional reward signals, including Format Rewards and LLM-as-a-Judge evaluations.

\subsection{\textit{Trainer Layer}: Policy Optimization}

The \textit{Trainer Layer} serves as the core optimization engine, dedicated to fine-tuning the agents' memory management policies. Currently, we implement a trainer based on the Group Relative Policy Optimization (GRPO) algorithm \cite{shao2024GRPOdeepseekmathpushinglimitsmathematical}. During the training loop, GRPO samples a group of rollout trajectories for each input and updates the policy based on relative advantages computed from the environment's rewards. Furthermore, this layer supports comprehensive hyperparameter configurations and natively integrates SwanLab \cite{Zeyilin_SwanLab_2023}. This allows researchers to monitor training metrics, reward distributions, and generation trajectories in real time, ensuring a transparent and customizable reinforcement learning process.

\section{Empirical Study}

To demonstrate the modularity of \textbf{MemFactory}, we assemble three representative agents inspired by classic memory paradigms: \texttt{MemoryR1Agent} based on Memory-R1 \cite{yan2026memoryr1}, \texttt{MemoryAgent} based on MemAgent \cite{yu2025memagent}, and \texttt{MemoryRMMAgent} based on RMM \cite{tan2025prospectretrospectreflectivememory}. These three agents are provided out-of-the-box for immediate use. 

To validate the effectiveness of the framework, we conduct our empirical study specifically on \texttt{MemoryAgent}. The original MemAgent work provides a comprehensive, publicly available dataset for both training and evaluation, making it the ideal choice for our experimental design.

\subsection{Experimental Setup}

We utilize the training and evaluation datasets provided by the MemAgent release \cite{yu2025memagent}. We train MemAgent-style agents using two open-source Large Language Models (LLMs) as base models: \texttt{Qwen3-1.7B} and \texttt{Qwen3-4B-Instruct} \cite{yang2025qwen3technicalreport}. The training data are adapted from the original MemAgent training dataset. To improve training efficiency while preserving the long-context nature of the tasks, we simplify the training samples by reducing the context length to approximately one-third of the original, resulting in 50 to 80 documents per sample.

For evaluation, we select three specific test sets from the MemAgent data: two main-task datasets (\texttt{eval\_50} and \texttt{eval\_100}) and one out-of-distribution (OOD) dataset (\texttt{eval\_fwe\_16384}). Each model is trained for 250 valid steps. Detailed hyperparameter configurations are available in our open-source repository. To account for generation variance, reported scores are averaged over 4 independent trials per question (avg@4). Notably, the entire training and evaluation pipeline can be executed on a single NVIDIA A800 80GB GPU, demonstrating that MemFactory is highly reproducible even with modest computational resources.

\begin{table}[t]
    \centering
    \caption{Performance of \texttt{MemoryAgent} trained via MemFactory on three test sets (all scores averaged over 4 independent trials, denoted as avg@4).}
    \label{tab:memfactory_memagent}
    \resizebox{0.9\linewidth}{!}{
    \begin{tabular}{llcccc}
        \toprule
        \textbf{Model} & \textbf{Setting} & \textbf{eval\_50} & \textbf{eval\_100} & \textbf{eval\_fwe\_16384} & \textbf{Average} \\
        \midrule
        \multirow{2}{*}{\texttt{Qwen3-1.7B}}
        & Base checkpoint & 0.4727 & 0.4297 & \textbf{0.0332} & 0.3118 \\
        & + MemFactory RL & \textbf{0.5684} & \textbf{0.4863} & 0.0195 & \textbf{0.3581} \\
        \midrule
        \multirow{2}{*}{\texttt{Qwen3-4B-Instruct}}
        & Base checkpoint & 0.6523 & 0.5645 & 0.6270 & 0.6146 \\
        & + MemFactory RL & \textbf{0.7051} & \textbf{0.6309} & \textbf{0.6426} & \textbf{0.6595} \\
        \bottomrule
    \end{tabular}
    }
\end{table}

\subsection{Results and Analysis}

As shown in Table \ref{tab:memfactory_memagent}, MemFactory consistently improves the average performance for both base models. For \texttt{Qwen3-1.7B}, the framework achieves a relative increase of 14.8\% in the average score. This is driven primarily by strong gains in the main-task settings, although performance on the OOD benchmark slightly decreases. In contrast, the larger \texttt{Qwen3-4B-Instruct} model demonstrates more robust and uniform improvements across all evaluation sets. It achieves a 7.3\% relative increase on average, including consistent gains on the OOD benchmark, indicating that the larger model effectively transfers the learned recurrent memory policy to unseen settings.

In general, these results validate two key conclusions. First, MemFactory successfully reproduces and optimizes MemAgent-style recurrent memory policies using standard training datasets. Second, even with a lightweight single-GPU setup and simplified training data, the framework delivers measurable performance improvements on challenging long-context tasks. These findings confirm that MemFactory serves as a concise, user-friendly, and effective infrastructure for empirical Memory-RL research.

\section{Conclusion}

In this paper, we present \textbf{MemFactory}, a modular framework for memory-augmented agents. By abstracting the memory lifecycle into standardized operations across four core layers (\textit{Module}, \textit{Agent}, \textit{Environment}, and \textit{Trainer}), MemFactory provides an accessible foundation that simplifies the design, training, and innovation of memory architectures. Additionally, we assemble out-of-the-box agents inspired by classic works—such as Memory-R1 \cite{yan2026memoryr1}, MemAgent \cite{yu2025memagent}, and RMM \cite{tan2025prospectretrospectreflectivememory}—for immediate use.

Our empirical study confirms that MemFactory serves as a concise, user-friendly, and effective framework for Memory-RL research. While the current version still covers only a limited set of representative paradigms and leaves room for further improvement in training efficiency, these limitations do not diminish its practical value as a unified infrastructure for developing memory-augmented agents. We hope MemFactory can serve as a solid foundation for future research on modular memory systems and reinforcement learning for long-horizon agents.

\clearpage

\bibliographystyle{plainnat}
\bibliography{main}

\clearpage


\end{document}